\documentclass[a4paper]{article}
\usepackage{subfigure}
\usepackage{xcolor}
\usepackage{amsmath}

\usepackage{INTERSPEECH_v2}

\usepackage{hyperref}
\pdfoutput=1

\title{Quantized-Dialog Language Model for Goal-Oriented Conversational Systems}
\name{R. Chulaka Gunasekara, David Nahamoo, Lazaros C. Polymenakos,  
\\ Jatin Ganhotra, and Kshitij P. Fadnis}
\address{IBM Thomas J. Watson Research Center, Yorktown Heights, NY, USA}
\email{chulaka.gunasekara@ibm.com, nahamoo@us.ibm.com, lcpolyme@us.ibm.com, jatinganhotra@us.ibm.com, kpfadnis@us.ibm.com   
}

\begin{document}

\maketitle
\begin{abstract}
We propose a novel methodology to address dialog learning in the context of goal-oriented conversational systems.
The key idea is to quantize the dialog space into clusters and create a language model across the clusters, thus allowing for an accurate choice of the next utterance in the conversation. 
This quantized-dialog language model methodology has been applied to the end-to-end goal-oriented track of 
the latest Dialog System Technology Challenges (DSTC6). The objective is to find the correct system utterance from a pool of candidates in order to complete a dialog between a user 
and an automated restaurant-reservation system.
Our results show that the technique proposed in this paper achieves high accuracy regarding selection of the correct candidate utterance, and outperforms other state-of-the-art approaches based on neural networks. 


\end{abstract}
\noindent\textbf{Index Terms}: language models, goal-oriented conversational systems, dialog learning, deep learning for dialog
\section{Introduction}
The Dialog State Tracking Challenge (DSTC; now rebranded as Dialog System Technology Challenges) initiative was 
originally conceived to provide a benchmark for the evaluation of  dialog-management \mbox{systems. The latest of these challenges}, DSTC6 \mbox(\url{http://workshop.colips.org/dstc6/}), is divided into three different tracks, and our work addresses the end-to-end goal-oriented dialog track. This track focuses on a restaurant-reservation problem: the objective is to book a table in a restaurant satisfying a number of requirements given by the user.

It is interesting to note that, as suggested in \cite{DBLP:journals/corr/BordesW16,DBLP:journals/corr/WilliamsAZ17}, methodologies based on rules may solve the goal-oriented dialog problem proposed in DSTC6 with full accuracy (i.e., no errors at all). By contrast, data-driven conversational systems (see, e.g., \cite{wang-lemon:2013:SIGDIAL,DBLP:journals/corr/ShangLL15,DBLP:journals/pieee/YoungGTW13}), are typically easier to apply to new domains and often perform in a satisfactory manner, but usually are less accurate than rule-based methods. 

In the implicit approach of developing dialog systems, the models are learned from the available interaction data, without using the traditional feature engineering and conversational components such as natural language understanding (NLU), dialog management, semantic matching etc. These models have the capability to continuously learn from the data and improve even after deployment, without cost and time consuming redesign of the features of individual components. Recent approaches of implicit dialog systems rely on applying deep learning techniques (e.g. sequence-to-sequence architectures) on the dialog data for prediction or generation of the next utterance in a dialog \cite{sutskever2014sequence, li2016deep, serban2016building}. But, the accuracies of these models on a variety of dialog datasets have been low \cite{lowe2017training}. 

During prediction, the current approaches select the next utterance from all possible responses that can be given at any turn, along with all their possible syntactic variations. This explodes the space of possible choices and leads to poor performance. To alleviate this problem, we propose the quantized representation of dialogs, which reduces the state space for prediction and improves the performance of data-driven conversational systems.

The paper is structured as follows. In the next section, we briefly describe the problem and dataset considered. Then, in Section \ref{sec:QM} we introduce the quantized-dialog language model and, based on it, we develop a next utterance selection method for our goal-oriented dialog. In Section \ref{sec:res}, we present results comparing the new approach with two reference schemes that rely on neural networks. Finally, we conclude our work with a few observations and indicate directions of future work.
\section{Problem statement and dataset}
\label{sec:PROBL}

The main problem in the goal-oriented dialog-learning track in DSTC6 is further divided into four major subtasks: i) Issuing API calls, ii) Updating API calls, iii) Displaying options and iv) Providing extra information. The first two subtasks address dialog interaction for the collection of user requirements for restaurant reservation: atmosphere, location, cuisine, number of people in the party, and price range. Once these five requirements have been fulfilled, an API call is issued in order to retrieve (at least three) restaurants from a knowledge database that satisfy the requirements. Subtask three is about presenting the user with possible restaurant choices and subtask four is about providing additional information related to the restaurant option the user selected (i.e., restaurant address and phone number). There is a fifth subtask that combines the four subtasks mentioned and aims at modeling a complete dialog system for restaurant reservation.

The training data released in the challenge consists of 10,000 dialogs per subtask (including the fifth task). The automated dialog system evaluated as part of the challenge has to select the correct next utterance in each dialog subtask out of a number of possible candidates. Test data consists of four test sets, each with 5,000 dialogs split equally (1000 dialogs) over the five tasks.
These four sets will be referred to in this paper as tst1, tst2, tst3 and tst4.
The first test set is based on the same type of dialogs that occur in the training data.
The second set incorporates out-of-vocabulary entities (e.g., new cuisine types) and restaurants extracted from a knowledge database that is different from the one used in the training data.
The third test set includes six entity types in the user requirements (the additional entity type is related to dietary restrictions) whereas there are only five in the training data.
Out-of-vocabulary entities corresponding to the six types as well as the restaurants from the new knowledge database are combined in the fourth set.
\section{A quantized language model for goal-oriented dialog learning}
\label{sec:QM}
\begin{figure*}[tbp]
\centering

        \subfigure[Pipeline of the training process]{
        
                \includegraphics[scale=0.26]{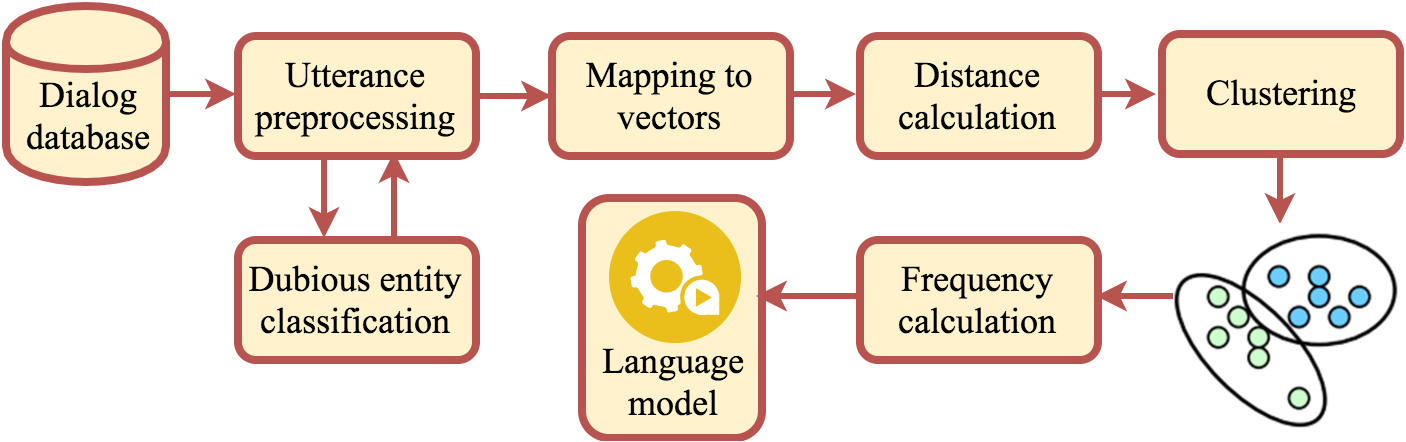}                \label{fig:training}
          }\\\vspace{0.5cm}

        \subfigure[Pipeline of the runtime process]{
                \includegraphics[scale=0.26]{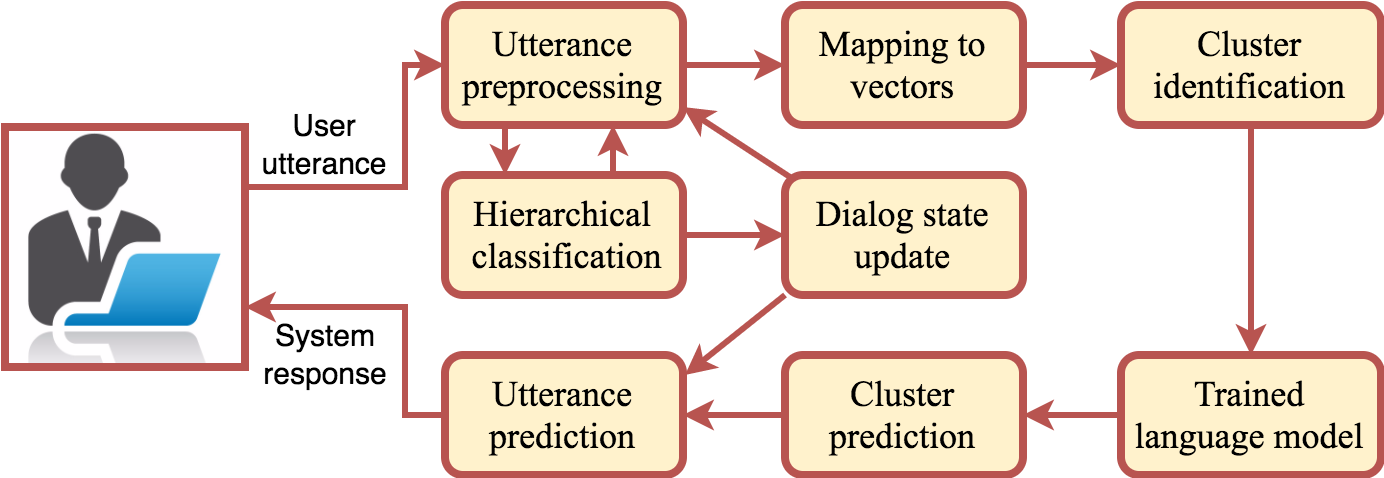}
                \label{fig:testing}
          }
\caption{The training and runtime processes of the quantized-dialog language model.}
\label{fig:training_runtime}
\end{figure*}

In this section we discuss our main contribution, which is the \textit{quantized-dialog language model (QDLM)}. In Figure \ref{fig:training_runtime} we show the model components for training and prediction (runtime), that we discuss in the sequel.

\subsection{Utterance preprocessing}\label{preprocess}
Natural language utterances usually contain information that may be unnecessary in a decision-making process. 
To remove irrelevant information, we preprocess the utterances via delexicalization of entities (or more generally parameters) relevant 
to the dialog goal. 
User utterances that contain entity values found in the knowledge base are transformed into a corresponding vector of entity
types. During this transformation, the entity types that were identified in previous utterances are also preserved.
The order of the entity types in the vector is: cuisine\textunderscore type, location, num\textunderscore people, price\textunderscore range and atmosphere.

An example of preprocessed dialog is shown in Table~\ref{tab:pre_proc}. 
We also detect dubious statements regarding entity/parameter selection by the user 
(see, e.g., utterance $11$ in that table; more details will be discussed later in Section \ref{ds_update}).
Constructing the language model with this delexicalized representation helps the quantized language model 
learn the dialog policy more accurately.

In Task~$3$ and~$4$, the set of restaurants recommended (obtained by means of the API call) is also preprocessed to 
identify the order in which the restaurants should be presented to the user. 
Consistent with our delexicalization approach, the name of the restaurant which should be proposed in the $i^{th}$ place is modified 
to \texttt{RESTAURANT\textunderscore i\textunderscore NAME}, and each of its properties (e.g., location, price, rating, etc.)
are accordingly changed to \texttt{RESTAURANT\textunderscore i\textunderscore PROPERTY}.

\subsection{Utterance quantization and language model}\label{lm_constuct}

Once the preprocessing is completed, the utterances are embedded into a vector space representation using the Bag-of-Words encoding \cite{clark2003neural} scheme. 
More complex procedures such as the Skip-Thought algorithm (see, e.g., \cite{kiros2015skip}) can also be considered 
for this process. 
Next, the utterances are clustered in the embedding space to create a quantized representation. 
In this implementation, each different vector was identified with a different cluster. Once the clusters are identified, each utterance can be represented by the identifier associated with the cluster that particular utterance belongs to. This yields to a quantized representation of utterances. 
Once the utterances are quantized, a conversation can be represented as a sequence of clusters. 
For example, consider a dialog $D$, which comprises the sequence of utterances $\{u_1, u_2, ..., u_N\}$, 
where $u_i$ is a natural language utterance.
Following the quantization, the dialog $D$ can be represented as a set of numbers $\{c_1, c_2, ..., c_N\}$, 
where each~$c_i$ corresponds to the cluster identifier where $u_i$ belongs. 

One of the simplest methods to assign probabilities to sequences of tokens is the 
n-gram language model~\cite{brown1992class}. 
%
The high-level idea is that the probability values for the candidate utterances are calculated using an n-gram language model constructed on the clusters.
In constructing the language model, we consider cluster transitions in all dialogs to calculate the transition probabilities between the clusters. 
This n-gram language model estimates the probability $P(c_i | c_{i-n}, ... , c_{i-1})$.

\begin{table*}[!ht]
\begin{center}
\caption{Example of dialog from Task 1 with the original utterances, the respective preprocessed representations and the dialog state constructed during the runtime.}
 \label{tab:pre_proc}
   \begin{tabular}{| c | p{5cm} | p{5cm} |p{5cm}|}
     \hline
     \textbf{Turn} & \textbf{Original Utterance} & \textbf{Preprocessed representation}  & \textbf{Dialog state}\\ \hline
     1 & \texttt{hello} & \texttt{hello} & \texttt{\{\}} \\ \hline
     2 & \texttt{hello what can i help you with today} & \texttt{hello what can i help you with today} &  \texttt{\{\}} \\ \hline
     3 & \texttt{i'd like to book \hspace{40pt} a table with \hspace{40pt}  a business atmosphere with spanish cuisine} & \texttt{CUISINE\textunderscore TYPE \newline NONE \newline NONE \newline NONE \newline ATMOSPHERE} &  \texttt{CUISINE\textunderscore TYPE\hspace{3pt}: spanish, ATMOSPHERE\hspace{11pt}: business}\\ \hline
     4 & \texttt{i am on it} & \texttt{i am on it} &  \texttt{CUISINE\textunderscore TYPE\hspace{3pt}: spanish, ATMOSPHERE\hspace{11pt}: business}\\ \hline
     5 & \texttt{\textless silence\textgreater} & \texttt{silence} &  \texttt{CUISINE\textunderscore TYPE\hspace{3pt}: spanish, ATMOSPHERE\hspace{11pt}: business} \\ \hline
     6 & \texttt{where should it be} & \texttt{where should it be} &  \texttt{CUISINE\textunderscore TYPE\hspace{3pt}: spanish, ATMOSPHERE\hspace{11pt}: business}\\ \hline
     7 & \texttt{find me one in bombay, paris will be too complicated} &  \texttt{CUISINE\textunderscore TYPE \newline LOCATION \newline NONE \newline NONE \newline ATMOSPHERE}  &  \texttt{CUISINE\textunderscore TYPE\hspace{3pt}: spanish, ATMOSPHERE\hspace{11pt}: business, LOCATION\hspace{22pt}: bombay}\\ \hline
     8 & \texttt{how many people would be in your party} &  \texttt{how many people would be in your party} &  \texttt{CUISINE\textunderscore TYPE\hspace{3pt}: spanish, ATMOSPHERE\hspace{11pt}: business, LOCATION\hspace{22pt}: bombay} \\ \hline
     9 & \texttt{for eight people please} &  \texttt{CUISINE\textunderscore TYPE \newline LOCATION \newline NUM\textunderscore PEOPLE \newline NONE \newline ATMOSPHERE} &  \texttt{CUISINE\textunderscore TYPE\hspace{3pt}: spanish, ATMOSPHERE\hspace{11pt}: business, LOCATION\hspace{22pt}: bombay, NUM\textunderscore PEOPLE\hspace{13pt}: eight} \\ \hline
     10 & \texttt{which price range are you looking for} &  \texttt{which price range are you looking for} &  \texttt{CUISINE\textunderscore TYPE\hspace{3pt}: spanish, ATMOSPHERE\hspace{11pt}: business, LOCATION\hspace{22pt}: bombay, NUM\textunderscore PEOPLE\hspace{13pt}: eight} \\ \hline
     11 & \texttt{expensive is tempting but cheap may be more reasonable} &  \texttt{CUISINE\textunderscore TYPE \hspace{50pt} LOCATION \hspace{50pt} NUM\textunderscore PEOPLE \hspace{50pt} DUBIOUS \hspace{50pt} ATMOSPHERE} &  \texttt{CUISINE\textunderscore TYPE\hspace{3pt}: spanish, ATMOSPHERE\hspace{11pt}: business, LOCATION\hspace{22pt}: bombay, NUM\textunderscore PEOPLE\hspace{13pt}: eight} \\ \hline
     12 & \texttt{whenever you're ready} &  \texttt{whenever you're ready}  &  \texttt{CUISINE\textunderscore TYPE\hspace{3pt}: spanish, ATMOSPHERE\hspace{11pt}: business, LOCATION\hspace{22pt}: bombay, NUM\textunderscore PEOPLE\hspace{13pt}: eight}\\ \hline
     13 & \texttt{let's do moderate price range, and keep expensive price range for another day} &  \texttt{CUISINE\textunderscore TYPE \hspace{50pt} LOCATION \hspace{50pt}  NUM\textunderscore PEOPLE \hspace{50pt} PRICE\textunderscore RANGE \hspace{50pt} ATMOSPHERE} &  \texttt{CUISINE\textunderscore TYPE\hspace{3pt}: spanish, ATMOSPHERE\hspace{11pt}: business, LOCATION\hspace{22pt}: bombay, NUM\textunderscore PEOPLE\hspace{13pt}: eight, PRICE\textunderscore RANGE\hspace{8pt}: moderate} \\ \hline
     
     \hline
   \end{tabular}
 \end{center}

\end{table*}

\subsection{Runtime and utterance prediction}
The utterance prediction problem can be formalized as $\mathrm{arg}\,\mathrm{max}_u \, P(u | u_1, ... u_{i-1})$, 
where $u$ denotes the utterance that maximizes the conditional probability with respect to all previous utterances.
In the quantized dialog space, this problem is transformed into the cluster prediction 
problem $\mathrm{arg}\,\mathrm{max}_c \, P(c | c_1, ... c_{i-1})$, where $c$ is the cluster identifier that 
maximizes the conditional probability with respect to the clusters associated with all the previous utterances. 
Note that we can estimate this by the n-gram language model 
as follows: $$\mathrm{arg}\,\mathrm{max}_c \, P(c | c_1, ... c_{i-1}) \approx \mathrm{arg}\,\mathrm{max}_c \, P(c | c_{i-1}, ... c_{i-n}).$$ 
As the utterances within a cluster are similar to one another, any utterance in the predicted cluster $c$ 
can be selected as the predicted utterance $u$. In this implementation, previous $7$ clusters were used to determine the probability distribution of the next cluster using the language model.  

During the evaluation of all the candidates for a response, the cluster to which each candidate belongs is identified, and then the language model is used to calculate
the probability associated with that cluster. Then we rank the candidate responses using this probability value together with the constructed dialog state (discussed in Section \ref{ds_update}). 

\subsection{Dialog-state update}\label{ds_update}
\begin{figure*}[!t]
\centering
\includegraphics[width=0.6\textwidth]{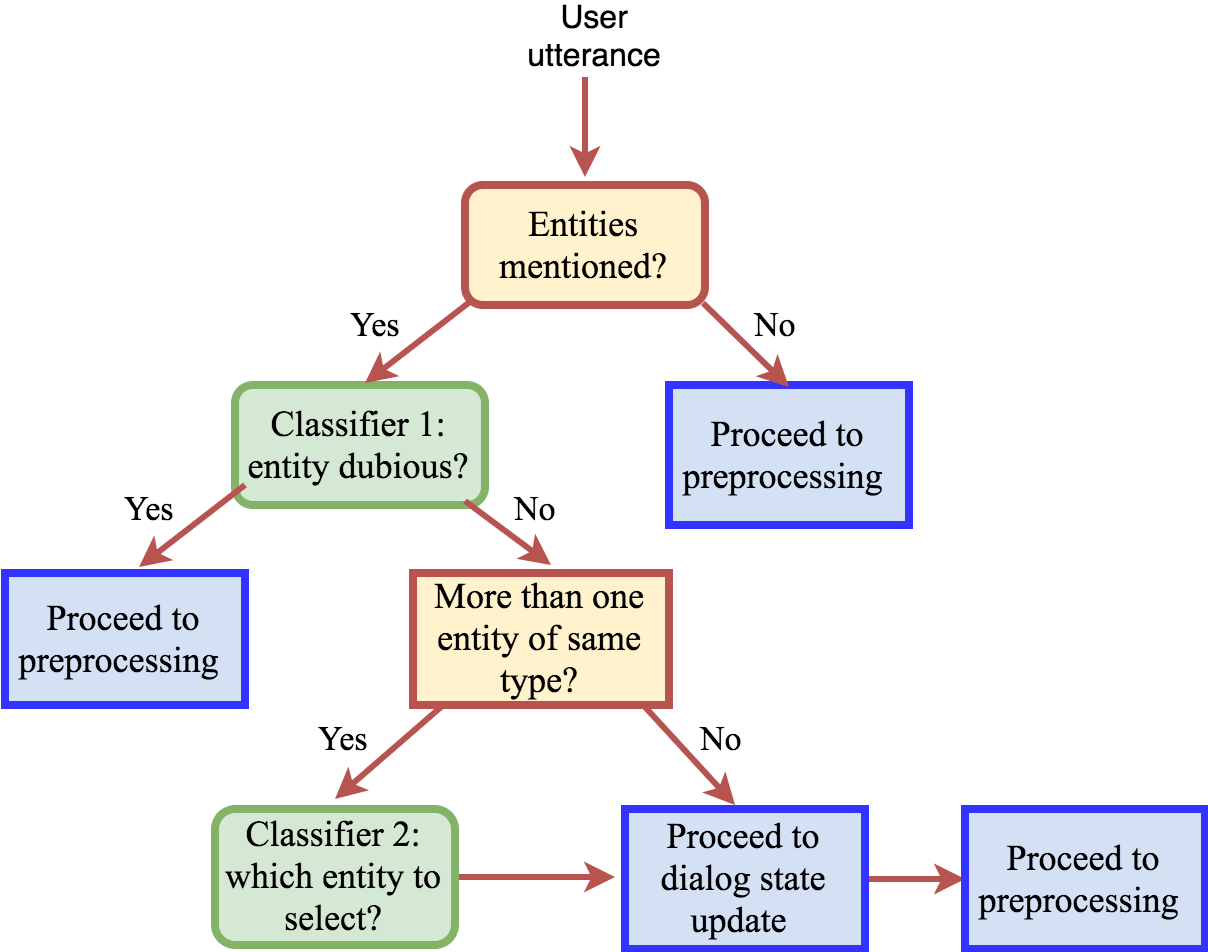}
\vspace{8pt}
\caption{Operation of the hierarchical classification process.}
\label{fig:ds_update}
\end{figure*}

In the approach presented in this paper, the dialog state corresponds to a key-value store associated with each 
entity specified by the user and its respective value (e.g., \texttt{CUISINE\textunderscore TYPE} and \texttt{indian}). 
The dialog-state update is based on a keyword matching algorithm, assisted by a hierarchical sentence classifier. 
The keyword matcher compares each word mentioned in a user utterance against all the entities in the knowledge base. 
Figure \ref{fig:ds_update} illustrates the operation of the dialog-state update module, and the last column of Table~\ref{tab:pre_proc} refers to the corresponding dialog state maintained through the dialog.
The hierarchical classifier is implemented using a publicly available text classification service\footnote{https://www.ibm.com/watson/services/natural-language-classifier/.}, and the operations of individual classifiers are explained below.


When an entity is detected in a user utterance, the first classifier (denoted as Classifier~1 in Figure~\ref{fig:ds_update}) 
 confirms the selection of that entity as a user specification. 
For example, the entity associated with the \texttt{CUISINE\textunderscore TYPE} is not confirmed as part of a user request in 
\texttt{one minute please,} \texttt{i am asking my friend} \texttt{if} \texttt{she} \texttt{wants to do} \texttt{spanish,}~\texttt{let's see}; by contrast \texttt{italian} is a confirmed requirement in the user utterance \texttt{wait, i am asking my friend and she wants italian, let's do that}.
Utterances with detected but unconfirmed entities are labeled as dubious. When preparing training data for this classifier, we label as  \texttt{dubious} the  user utterances with entities detected, followed
by the system utterance `whenever you’re ready', and we label as \texttt{not dubious} the other user utterances with just the entities detected. 

If two or more entities of the same type are identified in the same utterance but only one indicates user requirements, 
the second classifier (denoted as Classifier~2 in Figure~\ref{fig:ds_update})~allows selection of \mbox{the~correct~entity~in}
order to update the dialog state.
The training data for this classifier were prepared by, first, transforming utterances such as \texttt{let's do moderate price range,} \texttt{and keep expensive}~\texttt{price} 
\texttt{range for another day} 
into the form \texttt{let's do ENTITY\textunderscore 1 price range,} 
\texttt{and keep ENTITY\textunderscore 2 price range for another day}, and then identifying the correct entity \texttt{ENTITY\textunderscore 1} (label) from the subsequent API call.

Once the language model predicts that the next utterance is an API call, the key-value pairs associated with the current 
dialog state are used to fill in the slots of the API call. 
The API call thus constructed is compared against the given candidates using (word-level) Levenshtein distance (see, e.g., \cite{heeringa2004measuring}) in order to rank them.

It is important to note that, because QDLM has been trained on delexicalized data, out-of-vocabulary entities and 
restaurants from new knowledge databases can be handled without modifying the approach - with the caveat that the restaurants need to have the same information fields as in the database used for training.
We stress that the only modification we introduced in order to deal with additional parameters in the user requirements was 
to increase accordingly the number of key-value pairs in the dialog state (i.e., six pairs were considered 
in two of the test sets).
Therefore, when QDLM predicts that the next utterance should be an API call, the number of slots 
necessary for the call are directly given by the existing number of key-value pairs in the dialog state.


\section{Experiments and results}
\label{sec:res}
In this section, we first present the two neural network approaches that were used for comparison with QDLM.
Then, we report and briefly analyze the results of the three techniques on the DSTC6 datasets.

\subsection{Baselines considered}

For our first baseline, we choose a relatively simple architecture: a multi-layer feed-forward neural network (\cite{nn01,nn02})\footnote{For this baseline approach we would like to acknowledge David Echeverr\'ia Ciaurri who contributed to its design and implementation.}.
This network is applied to each dialog and candidate answer to yield a confidence metric defined in the interval $[0,1]$.
In all cases, a logistic function is considered after the output layer.

The first two tasks in DSTC6 are addressed by means of a network with one hidden layer of size $100$ with rectified 
linear unit as activation function.
The networks for Task 3 and 4 have three and two hidden layers, respectively, all of them of size $100$ 
(and also followed by rectified linear units). These networks were optimized, for each task individually, for accuracy and convergence rate by varying number of hidden layers, hidden layer sizes and activation layer functions.
The dialogs in Task 5 were first classified into four subtasks and then four networks of the same type as those were used to predict the next candidate utterance. 
%
Binary cross-entropy was minimized for each network by means of the Adam optimizer (see, e.g., \cite{optnn01}).
From all the dialogs in the dataset, 75\% of them are used to train the networks.
%

The input to each network was designed to allow ready application of the approach to other domains than restaurant 
reservations
(e.g., hotel-booking engines).
In all five tasks, the input vector includes the last four dialog utterances, each mapped to a $50$-dimensional vector.
In the first four tasks, the input incorporates the $50$-dimensional vector encoding of the candidate utterance as well. 
Additionally, Task 1 and 2 input encodes entities in the dialog and candidate answer by means of two $20$-dimensional vectors each corresponding to the $20$ different entities in the training data. For Task~3, we use instead  a vector with five components (the encodings for up to five names of restaurants that can be recommended 
by the system) and one scalar (the encoding for the name of the restaurant, if any, in the 
candidate utterance). Similarly, for Task 4 we use a vector with three components (the encodings of the name 
of the restaurant selected by the user and the corresponding address and phone number).
The current implementation for the neural-network classifier described here only handles five different types 
of entities (thus, it is not applicable to requests based on six or more types of entities). An additional limitation is discussed in ~\cite{vinyals2012revisiting}, where the feed-forward neural-network approach is shown to be unable to capture the functional dependency between a given output and past inputs in the network.

Our second baseline is based on end-to-end memory networks, recently proposed in~\cite{DBLP:journals/corr/BordesW16}. This approach has exhibited promising results on 
natural language processing tasks such as question answering (see, e.g., \cite{DBLP:journals/corr/WestonBCM15}).
Memory Networks \cite{DBLP:journals/corr/WestonCB14} operate by writing and then iteratively reading from memory. They compute multiple internal states using hops and the memory values are learned by soft-attention of user query with the previous dialog utterances. The internal state, combined with the context is later used to predict the required response. The concept of multiple computation hops allows to attend over different parts of context (previous dialog utterances) during each hop and hence captures an improved internal state for the overall dialog. In this implementation, the previous utterances and the user query were passed to the network encoded in bag-of-words representation, and the network learns the representations for the vocabulary, using cross-entropy loss objective function for the predictions.

An advantage of using end-to-end memory networks for the DSTC6 challenge is that they provide a common architecture for 
all five tasks and require less 
`feature engineering’ than the feed-forward neural network based classifier and QDLM approaches discussed above.
In these experiments, we split the training data into two parts, 80\% of it for training the network 
and the remaining 20\% for testing the network trained.

\subsection{Discussion of results}

\begin{table}[h!]
\begin{center}
\caption{Results for the training data obtained with the methods based on the quantized-dialog language model (QDLM), 
with the feed-forward neural network (FFNN) and with the memory network (MN).}
\label{tab:train}
\begin{tabular}{cccc}
\hline
   Task &     QDLM    &   FFNN           & MN               \\ \hline
	1 &   0.999   &   0.995 (0.995)  &   0.992 (0.997)  \\ \hline
	2 &   1.000   &   0.997 (0.996)  &   1.000 (1.000)  \\ \hline
	3 &   1.000   &   0.999 (0.998)  &   0.932 (1.000)  \\ \hline
	4 &   1.000   &   1.000 (0.999)  &   0.897 (1.000)  \\ \hline
	5 &   1.000   &   0.996 \textcolor{white}{(0.999)}    &                             0.920 (0.975) \\ \hline \hline
 Average  &   1.000   &   0.997 \textcolor{white}{(0.999)}    &                             0.948 \textcolor{white}{(0.999)} \\ \hline
\end{tabular}
\end{center}
\caption*{Notes: For each task and method, the first number is the accuracy for the testing subset of dialogs and
the second number (in parenthesis) is the accuracy for the training subset of dialogs. No accuracy is reported for QDLM 
regarding the training subset since this technique, unlike the other two approaches, does not explicitly 
perform a training iteration. The last row is the average accuracy over the five tasks in the dataset.}
%
%

\begin{center}
\caption{Results for the test data obtained with the quantized-dialog language model approach.}
\label{tab:test}
\begin{tabular}{ccccc}
\hline
 Task       &  tst1  &  tst2  &  tst3  & tst4    \\ \hline
  1         & 1.000  & 1.000  & 0.866  & 0.882   \\ \hline
  2         & 1.000  & 1.000  & 1.000  & 1.000   \\ \hline       
  3         & 0.998  & 0.997  & 1.000  & 1.000   \\ \hline
  4         & 1.000  & 1.000  & 1.000  & 1.000   \\ \hline
  5         & 0.986  & 0.985  & 0.957  & 0.953   \\ \hline \hline
  Average   & 0.997  & 0.996  & 0.965  & 0.967   \\ \hline
\end{tabular}
\end{center}
\caption*{Notes: A brief description of the four sets (tst1, tst2, tst3 and tst4) was given in Section \ref{sec:PROBL}. 
The accuracy obtained is given for each task and set. The last row is the average accuracy over the five tasks in each of these sets.}
\end{table}

Accuracy results for the training data (all five tasks) obtained with quantized-dialog language model (QDLM), feed-forward neural network (FFNN) and memory network (MN) are shown in Table~\ref{tab:train}. 
The last row in that table represents the average performance. We had to choose one method for submitting our results for the test data for the challenge. The decision for dropping our MN method was justified by the fact that the average accuracy for this method, i.e., 0.948, was clearly lower than the average accuracy of the other two techniques. Despite the fact that QDLM and FFNN show comparable performance for the training data, we observed that QDLM is better suited to handle additional parameters (as the ones present in the sets tst3 and tst4). Therefore, we only evaluated QDLM on the test data. 


The results obtained are summarized in Table~\ref{tab:test} for four test data tst1, tst2, tst3 and tst4, with an average accuracy of 0.981. The results for tst1 and tst2 are similar to what we obtained for the training set with accuracy values very close to one. We conclude that out-of-vocabulary entities and restaurants are handled successfully. QDLM shows lower accuracy in tst3 and tst4. While our technique can handle the API-call candidates with additional entity type, it does not properly deal with candidate utterances that query about the value of the additional entity type. We are working on a solution to this problem by further quantization of entity types.

 

\section{Conclusions and future work}


In this paper, we have introduced the Quantized-Dialog Language Model (QDLM) technique for constructing goal-oriented conversational systems. The method has been applied to the end-to-end goal-oriented dialog learning track of the sixth edition of the Dialog System Technology Challenges (DSTC6). The dataset in this track is a large collection of restaurant-reservation conversations. The challenge is constructed for the design of systems that can accurately select the correct utterance from a pool of candidate utterances that appropriately completes a given dialog. 

The QDLM technique achieved around 98\% candidate selection accuracy across 4 test sets each having 5000 conversations. Most of the errors were in test sets 3 and 4 where the candidate selection for dialogs with unseen entity types were not properly handled. Since the submission of the results for the challenge, we have made progress in this respect by further quantization in the candidate space for entity types to eliminate the dependency on multiple entity types. We have also started working on the integration of QDLM with sequence to sequence based natural language generation methods and we will report results in a future publication.

\bibliographystyle{IEEEtran}

\bibliography{main}

\end{document}